\documentclass[11pt,a4paper]{article}
\usepackage{times}
\usepackage{latexsym}
\usepackage{amsmath}
\usepackage{amsfonts}
\usepackage{graphicx}
\usepackage{caption}
\usepackage{setspace}
\usepackage[hyperref]{acl2021}
\usepackage{xcolor}
\usepackage{subfig}
\usepackage{float}
\usepackage{setspace}

\hypersetup{colorlinks=true,linkcolor=darkblue,citecolor=darkblue,filecolor=magenta,urlcolor=darkblue}

\usepackage{microtype}

\aclfinalcopy 

\title{{E}valuating {W}ord {E}mbeddings with {C}ategorical {M}odularity}

\author{S\'ilvia Casacuberta$^1$\thanks{\; Equal contribution} \,, \, Karina Halevy$^{1*}$, \, Dami\'an E. Blasi$^{1,2,3}$ \\\
  1: Harvard University 
  ~\;~\;~\; 2: MPI for Evolutionary Anthropology ~\;~\;~\; 3: HSE University \\\
  \texttt{\{scasacubertapuig, khalevy\}@college.harvard.edu} \\
  \texttt{dblasi@fas.harvard.edu}}

\date{}

\begin{document}
\maketitle
\begin{abstract}
We introduce categorical modularity, a novel low-resource intrinsic metric to evaluate word embedding quality. Categorical modularity is a graph modularity metric based on the $k$\nobreakdash-near\-est neighbor graph constructed with embedding vectors of words from a fixed set of semantic categories, in which the goal is to measure the proportion of words that have nearest neighbors within the same categories. We use a core set of 500 words belonging to 59 neurobiologically motivated semantic categories in 29 languages and analyze three word embedding models per language (FastText, MUSE, and subs2vec). We find moderate to strong positive correlations between categorical modularity and performance on the monolingual tasks of sentiment analysis and word similarity calculation and on the cross-lingual task of bilingual lexicon induction both to and from English. Overall, we suggest that categorical modularity provides non-trivial predictive information about downstream task performance, with breakdowns of correlations by model suggesting some meta-predictive properties about semantic information loss as well. 
\end{abstract}

\section{Introduction}\label{sec:intro}
Word embeddings represent words and phrases in continuous low-dimensional vector spaces. They are usually trained with neural language models or word collocations~\cite{bengio, collobert2, mikolovb, levy}, such that similar assignments in a space reflect similar usage patterns. The rise of monolingual embeddings such as Word2Vec~\cite{mikolovb}, GloVe~\cite{glove}, and FastText~\cite{fasttext1}, coupled with the need to transfer lexical knowledge across languages, has also led to the development of cross-lingual word embeddings, in which different languages share a single distributed representation and are mapped into the same vector space. Such methods use different  bilingual supervision signals (at the level of words, sentences, or documents) with varying levels of strength~\cite{ruder}.

A central task in the study of word embeddings is finding metrics to evaluate their quality. These metrics can either be \textit{extrinsic}, where embeddings are used as input features for downstream NLP tasks and evaluated on their performance, or \textit{intrinsic}, where embeddings are directly tested for how well they capture syntactic or semantic properties in their own right~\cite{qiu}. Extrinsic methods are not always feasible for low-resource languages due to a lack of annotated data. Moreover, downstream model components can be fine-tuned to achieve higher performance on certain tasks without necessarily indicating improvement in the semantic representation of words in an embedding space~\cite{leszczynski}.

This paper presents \textit{categorical modularity}, a low-resource intrinsic evaluation metric for both monolingual and cross-lingual word embeddings based on the notion of graph modularity. The underlying principle is that \textit{in good embeddings, words in the same semantic category should be closer to each other than to words in different categories}. We quantify this by building the $k$-nearest neighbor graph with a fixed set of words' semantic categories and computing the graph's modularity for a given embedding space. \textit{Modularity} measures the strength of division of a graph with densely connected groups of vertices, with sparser connections between groups~\cite{newman}.

We source our semantic categories from ~\citet{binder}. In contrast to other semantic and ontological categories in the literature, these have been motivated by a set of experiential attributes with neurobiological consistency, covering  sensory, motor, spatial, temporal, affective, social, and cognitive dimensions. We refer to these attributes collectively as \textit{Binder categories}. The resulting dataset consists of 500 English words, each labeled with three categories at three levels of semantic granularity. For example, the word \textit{chair} belongs to \textit{Concrete Objects} (Level~1), \textit{Artifacts} (Level~2), and \textit{Furniture} (Level~3). 442 words are pulled from Binder, on top of which we add a few words to even out distributions of categories and replace a few English-specific words with words that are more easily translated to non-English languages. 

We then translate these 500 English words into 28 more languages, selected based on their availability in the form of pre-trained vectors from the MUSE library~\cite{muse}. We produce 300-dimensional embeddings for these words using three popular embedding models: the monolingual FastText~\cite{fasttext1} and subs2vec~\cite{subs2vec} models and the cross-lingual MUSE~\cite{muse} model. Using these embeddings, we obtain the nearest-neighbor sets among the 500 words within each (language, model) pair and use those relationships to calculate a modularity score for the pair. We compare modularity scores to performance on three downstream tasks: sentiment analysis (monolingual classification), word similarity (monolingual regression), and word-level bilingual lexicon induction (BLI, cross-lingual regression) both to and from English. We obtain moderate to strong positive correlations on all three tasks, with slightly stronger results on the monolingual tasks. We also provide an analysis of correlations broken down by individual model and explore potential meta-predictive properties of categorical modularity.

We further show that estimating modularity on Binder categories yields relevant information that cannot simply be derived from naturally occurring distributions of word clusters in embedding spaces. We show this by replicating all three downstream task correlation analyses with modularity scores based on clusters obtained with unsupervised community detection methods~\cite{clauset}, which we henceforth refer to as \textit{unsupervised clusters}. 
After establishing the utility of categorical modularity, we show some of its use cases for comparing and selecting models for specific NLP problems, and we discuss preliminary results about the individual categories we find to be most predictive of downstream task performance. 

Our code and data are available to the public.\footnote{\url{https://github.com/enscma2/categorical-modularity}}

\section{Related Work}

\subsection{Word Embedding Evaluation Metrics}
While word embeddings have become crucial tools in NLP, there is still little consensus on how to best evaluate them. Evaluation methods commonly fall into two categories: those motivated by an \textit{extrinsic} downstream task and those motivated by the \textit{intrinsic} study of the nature of semantics and the cognitive sciences~\cite{survey}. Intrinsic and extrinsic metrics do not always align, as some models have high quality as suggested by intrinsic scores but low extrinsic performance, and vice versa~\cite{schnabel, glavas}.

Some commonly used methods of extrinsic evaluation include named entity recognition~\cite{collobert1} ---including the datasets \textit{CoNLL-2002} and \textit{CoNLL-2003}~\cite{tjong}---, sentiment analysis~\cite{schnabel}, semantic role labeling, and part-of-speech tagging~\cite{collobert1}. Intrinsic evaluation methods include word semantic similarity~\cite{baroni}, concept categorization~\cite{baroni}, and experiments on neural activation patterns~\cite{sogaard}.

Our categorical modularity metric is inspired by ~\citet{modularity}. They study the modularity of cross-lingual embeddings based on the premise that different languages are well-mixed in good cross-lingual embeddings and thus have low modularity with respect to language. Our metric improves upon the modularity proposed in ~\citet{modularity} by overcoming the problem caused by low modularity potentially occurring with a purely random distribution of word vectors and being mistaken for high embedding quality, as it is unlikely for a random distribution to coincidentally have highly modular clusters corresponding to Binder categories. Moreover, our metric is able to evaluate both monolingual and cross-lingual word embeddings and allow for comparisons between these types of embeddings (e.g., FastText and MUSE), and it incorporates cognitive information through the use of brain-based semantic categories. 

\subsection{Cognitive Approaches to NLP}
Recent work on word embeddings has explored the connections between NLP word representations and cognitively grounded representations of words. Such connections enrich both computational and neuroscientific research: external cognitive signals can enhance the capacity of artificial neural networks to understand language, while language processing in neural networks can shed light on how the human brain stores, categorizes, and processes words~\cite{muttenthaler}.

Cognitive approaches to lexical semantics propose a model in which words are defined by how they are organized in the brain~\cite{lakoff}. Based on this premise,~\citet{nora} propose \textit{CogniVal}, a framework for word embedding evaluation with cognitive language processing data. They evaluate six different word embeddings against a combination of 15 cognitive data sources acquired via eye-tracking, electroencephalography (EEG), and functional magnetic resonance imaging (fMRI). In a similar line of work, both~\citet{sogaard} and~\citet{beinborn} evaluate word embeddings using fMRI datasets.

The use of cognitive data in NLP goes well beyond the evaluation of word embeddings. ~\citet{utsumi} uses the neurosemantically inspired categories from~\citet{binder} to identify the knowledge encoded in word vectors. Among other conclusions, they find that the prediction accuracy of cognitive and social information is higher than that of perceptual and spatiotemporal information.

\section{Modularity and \boldmath$k$-NN Graphs}\label{sec:graphs}
The concept of modularity has origins in the field of network science, as first introduced by Newman~\cite{newman}. The goal of the modularity measure is to quantify the strength of the division of a network into clusters. Usually, such networks are represented with graphs. Intuitively, the modularity of a graph measures the difference between the fraction of edges in the graph that connect two nodes of the same category and the expected corresponding fraction if the graph's edges were distributed at random. Thus, the higher the proportion of edges between nodes that belong to the same category, the higher the modularity. 

In our case, we construct the pertinent graph with the $k$-nearest neighbors algorithm. Given a set $S_w$ of $N$ words and a set $S_c$ of categories such that each of the $N$ words belongs to exactly one of the categories in $S_c$, we map each of the $N$ words into a $d$-dimensional word embedding vector space and obtain a $d$-dimensional vector for each word. For each pair $(w_i, w_j)$, where $w_i, w_j \in S_w$ and $1 \leq i, j \leq |S_w|$, with corresponding $d$-dimensional vectors $v_i$ and $v_j$, we compute their cosine similarity (the cosine of the angle between them), which we denote by \texttt{similarity}$(i, j)$.

We create a matrix $M_D$ of dimensions $|S_w| \times |S_w|$, where entry $(M_D)_{i, j}$ is $\texttt{similarity}(i, j)$. For a given $k \in \mathbb{Z}_{>0}$, we build the $|S_w| \times |S_w|$ $k$-nearest neighbor matrix (denoted $k$-NNM) as follows: entry $(i, j)$ of $k$-NNM is equal to $1$ if and only if word $j$ is one of the $k$ nearest neighbors of word $i$ (i.e., if \texttt{similarity}$(i, j)$ is among the $k$ largest cosine similarities between $i$ and all other words in $S_w$). We note that $M_D$ and $k$\nobreakdash-NNM are not necessarily symmetric, as word $i$ being the $k$\nobreakdash-th nearest neighbor of word~$j$ does not imply the reverse. Finally, we define the $k$-NN graph of $S_w$ as the graph defined by $k$-NNM viewed as an adjacency matrix. We can now describe how to compute the modularity score following the schema in~\citet{modularity}.

Let $d_i$ denote the degree of node $i$, that is, $d_i = \sum_j (k$-NNM$)_{i, j}$, and let $g_i$ denote the category of word $i$. For each category $c \in S_c$, the expected number of edges within $c$ is
\begin{equation}
    a_c = \dfrac{1}{2m} \sum_i d_i \, I[g_i = c],
\end{equation}
where $m$ is the total number of edges in the $k$-NN graph and $I$ is the indicator function that evaluates to 1 if the argument is true and to 0 otherwise. 

The fraction of edges $e_c$ that connect words of the same semantic category $c$ is
\begin{equation}
    e_c = \dfrac{1}{2m} \sum_{i, j} (k\textrm{-NNM})_{i, j} \, I[g_i = c] \, I[g_j = c].
\end{equation}

By weighting the $|S_c|$ different semantic categories together, we calculate the overall modularity $Q$ as follows: 
\begin{equation}
    Q = \sum_{c = 1}^{|S_c|} (e_c - a_c^2).
\end{equation}
Finally, we normalize $Q$ by setting
\begin{equation}\label{eq:qnorm}
    Q_{max} = 1 - \sum_{c=1}^N a_c^2, \quad Q_{norm} = \dfrac{Q}{Q_{max}}.
\end{equation}
In our setting, $Q_{norm}$ indicates the modularity score of one (language, model) pair overall, but we denote by $Q_c$ the modularity of said (language, model) pair with respect to category $c \in S_c$. The definition of $Q_c$ (normalized) is deduced from Equation~\ref{eq:qnorm}:
\begin{equation}\label{eq:Qc}
    Q_c = \dfrac{e_c - a_c^2}{Q_{max}}.
\end{equation}

A higher value of $Q_{norm}$ indicates that a higher number of words that belong to the same categories appear connected in the $k$-NN graph. In Sections~\ref{sec:modularity} and \ref{sec:experiments}, we analyze the values $Q_{norm}$ for each of the languages, and in Section~\ref{sec:futurework}, we make some observations about the different values of $Q_c$.

\begin{figure}%
    \centering
    \subfloat[][\centering A high-modularity semantic $k$-NN graph.]{\includegraphics[width=6.2cm]{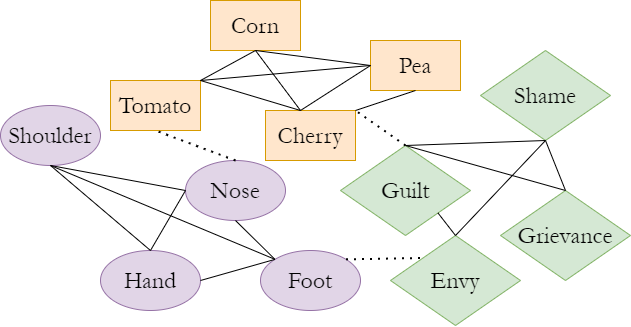}}%
    \qquad
    \subfloat[][\centering A low-modularity semantic $k$-NN graph.]{\includegraphics[width=5.6cm]{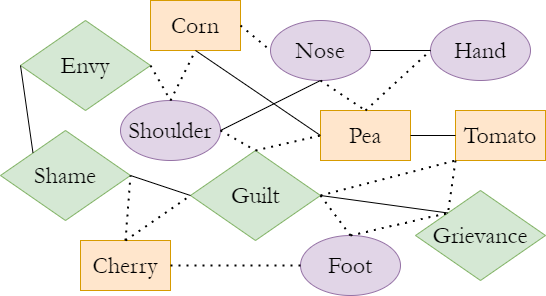}}%
    \caption{A toy visualization of the meaning of categorical modularity with some words pertaining to the Binder categories \textit{Body Parts} (purple ovals), \textit{Plants} (peach rectangles), and \textit{Emotions} (green rhombi). Dotted edges connect nodes of different categories, while solid edges connect nodes of the same category.}
    \label{fig:example}%
\end{figure}

In our conclusions about how our categorical modularity scores correlate with downstream task performance, we also want to prove that our selected neurosemantic-based categories are non-trivial and are better predictors than the unsupervised clusters that emerge from the embeddings. To find these clusters, we use the Clauset-Newman-Moore greedy modularity method~\cite{clauset}. This algorithm iteratively joins the pair of communities that most increases modularity until no such pair exists. For each value of $k$, we obtain the unsupervised communities in this manner and compute their modularity scores. In Section~\ref{sec:experiments}, we show that Binder categories are significantly better predictors than the unsupervised clusters using the same set of 500 words.

\section{Dataset}
In this section, we define the sets $S_N$ and $S_c$ of words and their semantic categories, respectively, that we use to compute categorical modularity scores for 29 languages.\footnote{The 29 languages are: Arabic, Bulgarian, Catalan, Croatian, Czech, Danish, Dutch, English, Estonian, Finnish, French, Greek, Hebrew, Hungarian, Indonesian, Italian, Macedonian, Norwegian, Polish, Portuguese, Romanian, Russian, Slovak, Slovenian, Spanish, Swedish, Turkish, Ukrainian, and Vietnamese.} As outlined in Section~\ref{sec:intro}, our motivation to take a cognitive approach in the study of word embeddings prompts us to use words and categories that reflect a brain-based computational model of semantic representation as in~\citet{binder}. We have 500 words (comprised of nouns, adjectives, and verbs) with 3 levels of categories, from most general (Level~1) to most specific (Level~3). Each word is tagged with 3 categories (one per level), which are listed in Table~\ref{table:binder}. After lifting 442 English words from Binder and adjusting the word set to optimize evenness of distribution across categories and translatability of concepts across languages, we manually translate the words to the 28 non-English languages\footnote{For languages with which we were not familiar, we solicited translations from colleagues, whom we compensated fairly for this work.} mentioned in footnote 2. 

\begin{table}[h!]
\centering
\begin{tabular}{ |p{7cm}| }
\hline
\setstretch{0.9}
{\footnotesize
\textbf{Level 1.} Concrete Objects, Concrete Events, Abstract Entities, 
Concrete Actions, Abstract Actions, States, Abstract Properties, Physical Properties.}
\\
\hline
\setstretch{0.9}
{\footnotesize
\textbf{Level 2.} Living Things, Other Natural Objects, Artifacts, 
Social Events, Nonverbal Sound Events, Weather Events, Miscellaneous,  Concrete Events, Abstract Constructs, Cognitive Entities, Emotions, 
Social Constructs, Time Periods, Body Actions, Locative Change Actions,
Social Actions, Miscellaneous Actions, Abstract Actions, 
States, Abstract Properties, Physical Properties.}
\\
\hline
\setstretch{0.9}
{\footnotesize
\textbf{Level 3.} Animals, Body Parts, Humans, Human Groups, Plants, 
Natural Scenes, Miscellaneous Natural Objects, Furniture, Hand Tools, Manufactured Foods, Musical Instruments, Places/Buildings, Vehicles, Miscellaneous Artifacts, Social Events, Nonverbal Sound Events, Weather, Events, Miscellaneous Concrete Events, Abstract Constructs, Cognitive Entities, Emotions, Social Constructs, Time Periods, Body Actions, Locative Change Actions, Social Actions, Miscellaneous Actions, Abstract Actions, States, Abstract Properties, Physical Properties.}
\\
\hline
\end{tabular}
\caption{Three levels of Binder categories.}
\label{table:binder}
\end{table}

\section{General Categorical Modularity}\label{sec:modularity}
\label{sec:categorical}
With the dataset of 500 words that belong to three levels of semantic categories, we compute the modularity scores of each of the 29 languages for each of the three word embedding models (which we refer to as 87 (language, model) pairs): FastText,\footnote{\url{https://fasttext.cc/docs/en/crawl-vectors.html}} MUSE,\footnote{\url{https://github.com/facebookresearch/MUSE\#download}} and subs2vec.\footnote{\url{https://github.com/jvparidon/subs2vec}} We briefly summarize the properties of each of these embeddings.

\textbf{FastText.} Monolingual embeddings for 157 languages trained on Common Crawl and Wikipedia that use CBOW with position-weights and character $n$-grams~\cite{fasttext1}. 

\textbf{MUSE.} Cross-lingual embeddings resulting from the alignment of 30 FastText embeddings into a common space under the supervision of ground-truth bilingual dictionaries~\cite{muse}.

\textbf{subs2vec.} Monolingual embeddings for 55 languages trained on the OpenSubtitles corpus of speech transcriptions from television shows and movies using the FastText implementation of the skipgram algorithm~\cite{subs2vec}. The authors claim that subtitles are closer to the human linguistic experience~\cite{subs2vec}. 

Information about the sizes of each (language, model) pair can be found in Appendix~\ref{app:b}. For each pair, we build the $k$-NN graph and compute modularity for different values of $k$ and different levels of categories, which we treat as our 2 hyperparameters. We consider small values for $k$ (namely $k \in \{2, 3, 4\}$) due to the fact that categories such as \textit{States} have as few as 4 words.

\section{Downstream Task Experiments}\label{sec:experiments}
We test the reliability of categorical modularity by running a few downstream tasks and computing the Spearman rank correlations between categorical modularity scores and performance on these tasks.

\begin{figure*}[h!]
  \centering
    \includegraphics[width=\linewidth]{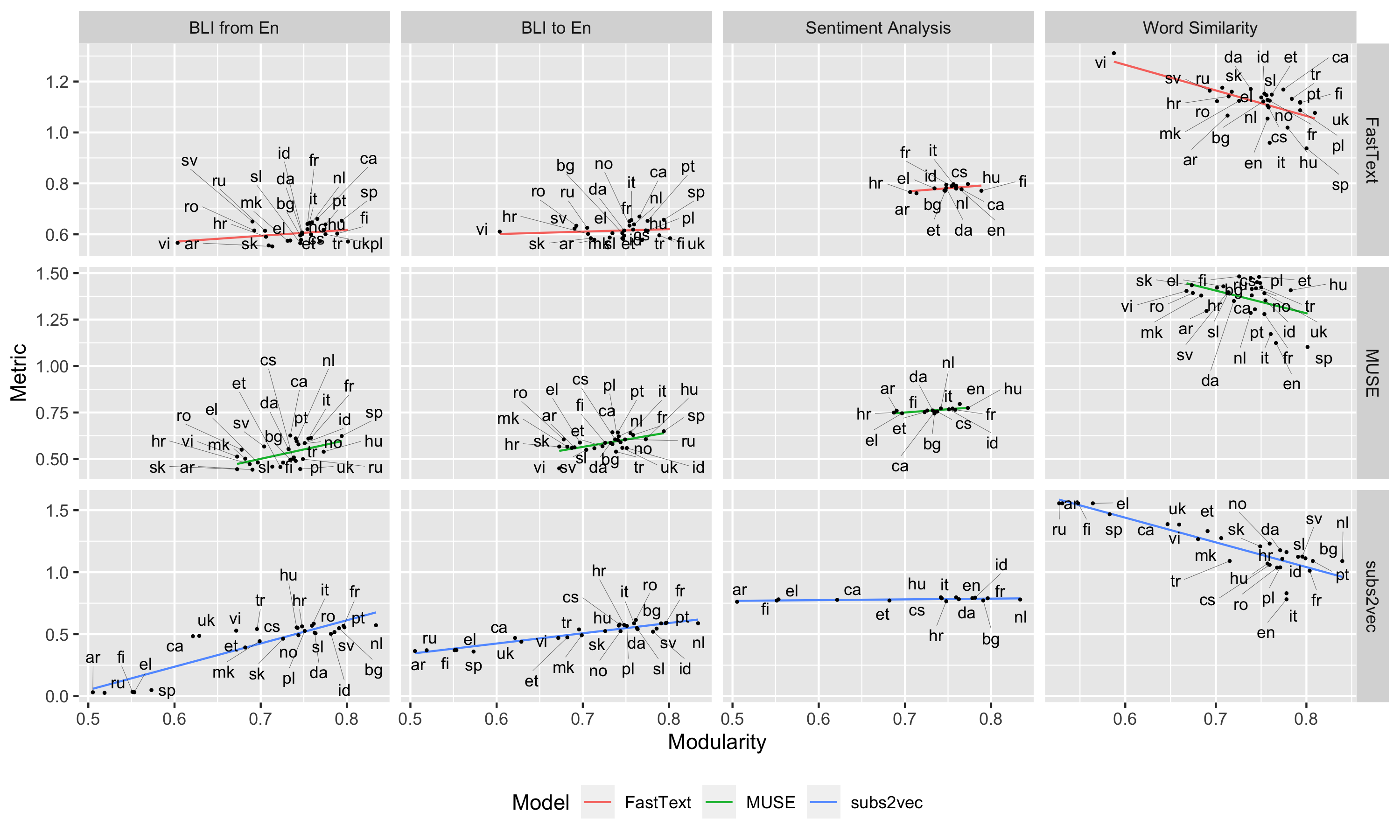}
    \caption{Summary of modularity vs performance metrics across tasks. Each language is represented with its 2-letter ISO 639-1 code. Hebrew, an outlier on the low end, is not included in this plot. Full modularity and performance data is included in our public GitHub repository.}
    \label{fig:main}
\end{figure*}

After determining the optimal set of hyperparameters ($k$ and level of semantic categories) for each task, we then compare the correlation produced by that set of hyperparameters with the correlation produced by the corresponding value of $k$ with the modularity of the unsupervised clusters constructed by the community detection algorithm described in Section~\ref{sec:graphs} to establish the non-triviality of the predictive properties of these chosen semantic categories. Table~\ref{table:spearman} provides a summary of correlation values for four tasks: movie review sentiment analysis (Sentiment), word similarity (WordSim), bilingual lexicon induction from English (BLI from), and bilingual lexicon induction to English (BLI to). Appendix~\ref{sec:appa} contains full tables with the correlation results. A visual summary of the results can be found in Figure \ref{fig:main}.

\begin{table}[h!]
\centering
\begin{tabular}{ccccc} 
\hline
\textbf{Task} & $\mathbf{\rho}$ & $\mathbf{\rho_{ft}}$ & $\mathbf{\rho_m}$ & $\mathbf{\rho_s}$ \\[0.05cm]
\hline
Sentiment & 0.54 & 0.44 & 0.68 & 0.46 \\
WordSim & 0.71 & 0.59 & 0.34 & 0.80 \\
BLI from & 0.55 & 0.40 & 0.54 & 0.76 \\
BLI to & 0.50 & 0.29 & 0.56 & 0.82 \\
\hline
\end{tabular}
\caption{Summary of Spearman correlations of categorical modularity with downstream task performance for Binder categories, aggregated across all models ($\mathbf{\rho}$) and broken down within FastText ($\mathbf{\rho_{ft}}$), MUSE ($\mathbf{\rho_m}$), and subs2vec ($\mathbf{\rho_s}$).}
\label{table:spearman}
\end{table}

\subsection{Sentiment Analysis}\label{sec:sentiment}
We first test our modularity scores through correlations with performance on the binary classification task of sentiment analysis, where the input is a movie review and the output is a binary label that corresponds to either positive or negative sentiment for that review. For this task, our data consists of $5{,}000$ randomly selected positive movie reviews and $5{,}000$ randomly selected negative reviews from the IMDB Movie Reviews dataset~\cite{imdb}. We randomly partition these $10{,}000$ reviews into 80\% training and 20\% testing data. Because this dataset is only available in English, we use the Google Translate API\footnote{\url{https://pypi.org/project/google-trans-new/}} to translate the data to 15 more languages (Arabic, Bulgarian, Catalan, Croatian, Czech, Danish, Dutch, Estonian, Finnish, French, Greek, Hebrew, Hungarian, Indonesian, and Italian) for a total of $n = 48$ observations. The languages and the dataset size of $10{,}000$ are chosen due to Google Translate API rate limits.

For each (language, model) pair, we convert the raw text of each review to a 300-dimensional embedding vector. We use the built-in black-box position-weighted continuous bag-of-words embedding model for FastText and subs2vec~\cite{fasttext1}, and we use a simple mean of individual word embeddings for MUSE, as the MUSE library does not have multi-word phrase embeddings built into its functionality. Using a vanilla linear support vector machine model with scikit learn's default settings,\footnote{\url{https://scikit-learn.org/stable/modules/svm.html}} we run the task on each language-model pair 30 times and record the mean accuracy and precision scores for each pair. We then calculate the Spearman correlations of each of the 9 modularity scores with both the accuracy and precision values. Furthermore, we analyze the overall merged correlations (taking all 48 data points for a given modularity score and performance metric) as well as the correlations within models (taking only the 16 data points within each single model), giving us a total of 72 Spearman correlation values.

We find that the optimal set of hyperparameters is Level 3 categories with $k = 2$, which gives a Spearman correlation of $\mathbf{\rho} = 0.54$ with the accuracy metric. Breaking it down by individual model, we have $\mathbf{\rho_{ft}} = 0.44$ for FastText, $\mathbf{\rho_m} = 0.68$ for MUSE, and $\mathbf{\rho_s} = 0.46$ for subs2vec. For $k = 2$, the correlations of unsupervised clusters with accuracy are $\mathbf{\rho} = 0.09$ for all 48 observations merged, $\mathbf{\rho_{ft}} = 0.1$, $\mathbf{\rho_m} = 0.4$, and $\mathbf{\rho_s} = 0.35$, providing evidence that Binder categories contain non-trivial predictive information that is not present in naturally emerging clusters.

\subsection{Word Similarity}\label{sec:wordsim}
Our next downstream task is the monolingual regression task of word similarity, in which the input is two words in one language and the output is a real number between 0 and 4 representing how similar the two words are (a higher score represents a greater degree of similarity). We use the English, Italian, and Spanish word pair datasets from SemEval-2017~\cite{camacho}, and we use the same Google Translate API from Section~\ref{sec:sentiment} to translate the English dataset into the remaining 26 languages. Each language's dataset then has 500 word pairs, which we randomly split into 400 training pairs and 100 testing pairs for each trial. 

Given a language and a model, we take each word pair, compute the 300-dimensional embeddings of both words, and calculate the Euclidean distance, Manhattan distance, and cosine similarity between the embeddings. We then feed these three scalars as a vector of inputs into a standard linear regression model from Python's scikit-learn package with default settings,\footnote{\url{https://scikit-learn.org/stable/modules/linear\_model.html}} whose output is the similarity score given in the dataset. To evaluate task performance, we compute the Mean Squared Error (MSE) loss for each run and record the mean MSE loss over 30 trials per (language, model) pair.

We then calculate the Spearman correlations of each of the 9 modularity scores with the negatives of the losses (such that positive correlation means that high modularity predicts high performance), both merged (87 data points) and within individual models (29 data points per model), for a total of 36 correlation values.

We find that all of the merged correlations are moderately to strongly positive. In particular, with the optimal hyperparameters of Level 2, $k = 2$, we have $\mathbf{\rho} = 0.71$ overall, $\mathbf{\rho_{ft}} = 0.59$ for FastText, $\mathbf{\rho_m} = 0.34$ for MUSE, and $\mathbf{\rho_s} = 0.8$ for subs2vec. In comparison, the correlations of the unsupervised cluster modularities with mean MSE loss for $k = 2$ are $\mathbf{\rho} = 0.27$, $\mathbf{\rho_{ft}} = 0.36$, $\mathbf{\rho_m} = 0.3$, and $\mathbf{\rho_s} = 0.42$, all weaker than their Binder counterparts.

\subsection{Word-Level Bilingual Lexicon Induction}
In addition to both monolingual classification and monolingual regression tasks, we also test our modularity metric on the cross-lingual regression task of bilingual lexicon induction. Using the ground-truth bilingual dictionaries provided by the publishers of MUSE~\cite{muse}, we run this task with the 28 non-English languages listed in footnote 2 in two directions: translation to and from English. We use the $5{,}000$-$1{,}500$ train-test split provided in the MUSE dictionary dataset and formulate the tasks as multivariate, multi-output regression tasks: for each observation in each (language, model) pair, we convert the English source word to its 300-dimensional embedding specified by the English version of the model and feed this vector as input to the same scikit-learn linear regression model as in Section~\ref{sec:wordsim}, of which the output is a 300-dimensional vector in the target language model space representing the embedding of the target word.

We follow this procedure in the other direction as well by converting source non-English words to embeddings in the appropriate non-English model spaces, feeding those embeddings into the linear regression model, and computing 300-dimensional predictions for the target English word vectors in the English model spaces. To measure task performance in the ``from English'' direction, we convert the ground-truth non-English target words into vectors in the corresponding non-English embedding model space, compute the cosine similarities between each ground-truth vector and its corresponding predicted vector, and record the mean of those cosine similarities as a measure of how close we are to the ground truth on average. We run 30 trials per (language, model) pair and record the mean of the mean cosine similarities.

In the ``to English'' direction, we similarly convert the ground-truth English target words into vectors and compute the mean cosine similarity over the prediction-ground-truth pairs. We calculate the Spearman correlations of each of the 9 modularity scores with the 30-trial means of mean cosine similarities in both directions. Once again, we calculate the correlations both across all models and within each individual model, yielding 72 total correlation values.

The optimal set of hyperparameters for the merged correlation in the ``from English'' direction is Level 3, $k = 2$, giving a moderate $\mathbf{\rho} = 0.55$ overall, $\mathbf{\rho_{ft}} = 0.4$ for FastText, $\mathbf{\rho_m} = 0.54$ for MUSE, and a strong $\mathbf{\rho_s} = 0.76$ for subs2vec. For comparison, the corresponding $k = 2$ correlations for unsupervised cluster modularities are $\mathbf{\rho} = 0.35$, $\mathbf{\rho_{ft}} = 0.04$, $\mathbf{\rho_m} = 0.27$, and $\mathbf{\rho_s} = 0.65$ --- all weaker than their Binder counterparts. 

The optimal set of hyperparameters for the merged correlation in the ``to English'' direction is also Level 3, $k = 2$, giving a moderate $\mathbf{\rho} = 0.5$ overall, a weak $\mathbf{\rho_{ft}} = 0.29$, a moderate $\mathbf{\rho_m} = 0.56$, and a very strong $\mathbf{\rho_s} = 0.82$. The corresponding unsupervised cluster correlations for $k = 2$ are $\mathbf{\rho} = 0.35$, $\mathbf{\rho_{ft}} = 0.04$, $\mathbf{\rho_m} = 0.27$, and $\mathbf{\rho_s} = 0.65$.

Notably, in both the word similarity and BLI tasks, $\mathbf{\rho_s}$ is significantly stronger than $\mathbf{\rho_{ft}}$ and $\mathbf{\rho_m}$. This may be due to the fact that compared to sources such as Wikipedia and Common Crawl, the subtitles used as training data for subs2vec are more representative of how the human brain semantically maps language, as suggested by the model's creators~\cite{subs2vec}.

Overall, these downstream task experiments suggest that categorical modularity is a non-trivially significant predictor of performance on both monolingual and cross-lingual NLP tasks (though it is stronger on monolingual tasks) and that it may have potential to be a meta-predictor of how well a particular model matches the information encoded in the human brain.

\section{Use Cases: Comparing and Selecting Models}
After having established substantial evidence of the predictive properties of categorical modularity, we present some examples of how the research community can make use of the metric for model evaluation and selection.

\subsection{Comparing Models within a Language}
The best hyperparameters for the tasks described in Section~\ref{sec:experiments} are Level 3 with $k = 2$ along with Level 2 with $k = 2$. Across the 29 languages at the latter, FastText has the highest modularity 9 times (Arabic, Catalan, Estonian, Finnish, Greek, Macedonian, Polish, Turkish, Ukrainian), while MUSE has the highest modularity 3 times (Hungarian, Russian, Spanish), and subs2vec has the highest modularity 17 times. For Level 3 with $k = 2$, FastText has the highest modularity 13 times, while MUSE has the highest modularity 2 times (Russian, Vietnamese), and subs2vec has the highest modularity 15 times. Though individual choices should be made with each language, this suggests that subs2vec may be a strong choice for monolingual tasks overall.

\subsection{Comparing Languages within a Model}
We also present some evidence that categorical modularity predicts bilingual lexicon induction performance moderately well, and predictive properties are especially strong within subs2vec. For the optimal set of hyperparameters found in that task within subs2vec (Level 2, $k = 2$, $\mathbf{\rho_s} = 0.77$ from English and $\mathbf{\rho_s} = 0.81$ to English), the languages with the highest modularities in subs2vec are Dutch ($0.84$), Portuguese ($0.81$), French ($0.80$), Bulgarian ($0.80$), Swedish ($0.80$), Indonesian ($0.79$), and English ($0.78$), while the languages with the lowest modularities are Catalan ($0.65$), Spanish ($0.58$), Hebrew ($0.58$), Greek ($0.56$), Finnish ($0.55$), Arabic ($0.53$), and Russian ($0.53$). This may suggest which languages have lower amounts of resources at this time and hence deserve more data collection efforts on the part of the NLP community, particularly within subs2vec's domain of subtitle and conversational data.

\subsection{Categorical Modularity as a Potential Meta-Predictor}
We find evidence that categorical modularity reveals some information about how well models map to the human brain, as suggested by subs2vec's significantly stronger correlations. This is particularly true with regression tasks. Given a new or existing embedding model, calculating its categorical modularities and assessing their correlations with regression tasks such as word similarity may reveal if the model space is representative of how linguistic information is encoded in the brain.

\section{Discussion and Future Work}\label{sec:futurework}
Categorical modularity shows promise as an intrinsic word embedding evaluation metric based on our preliminary experiments. We can envision extending this work in several directions. For one, we can calculate single-category modularities (denoted by $Q_c$ as defined in Equation~\ref{eq:Qc}) and test which individual categories contain the most predictive properties. Our limited experiments in this direction with the movie sentiment analysis task suggest that concrete and non-living categories have better predictive capabilities than abstract and living ones: for the sentiment analysis task, the 5 most strongly correlated categories are \textit{Nonverbal Sounds}, \textit{Artifacts}, \textit{Concrete Objects}, \textit{Vehicles}, and \textit{Manufactured Foods}, while the 5 least correlated categories are \textit{Abstract Properties}, \textit{Abstract Constructs}, \textit{Miscellaneous Actions}, \textit{Humans}, and \textit{Abstract Actions}.

We may also extend our work to more models and languages to see if the predictive properties truly hold across all languages and models. Additionally, as more multilingual research and data becomes available in this space, we may probe different sets of semantic categories, further downstream tasks (particularly multi-class classification, monolingual text generation, and sentence-level bilingual lexicon induction), and further variations of models used in downstream tasks (e.g., deeper neural networks instead of vanilla SVMs and linear regressions). We can also envision improvements upon the categorical modularity metric itself, perhaps by way of a lower-resource metric or a metric that works well on contextualized word embeddings for which the word-vector mappings may have more complex geometries. Our code and data, which are available to the public,\footnote{\url{https://github.com/enscma2/categorical-modularity}} can also enable researchers and practitioners to replicate our results and experiment with different models, words, languages, and categories.

\section{Conclusion}
In this paper, we introduce categorical modularity, a novel low-resource metric that may serve as a tool to evaluate word embeddings intrinsically. We present evidence that categorical modularity has strong non-trivial predictive properties with respect to overall monolingual task performance, moderate predictive properties with respect to cross-lingual task performance, and potential meta-predictive properties of model space similarity to cognitive encodings of language.

\section*{Acknowledgments}
This work was supported by the Radcliffe Institute for Advanced Study at Harvard University. SC and KH were supported by the School of Engineering and Applied Sciences at Harvard University under Prof. David Parkes. DEB acknowledges funding from the Branco Weiss Fellowship, administered by the ETH Zurich. DEB’s research was also executed within the framework of the HSE University Basic Research Program and funded by the Russian Academic Excellence Project ‘5-100’.


\section*{Impact Statement}

\subsection*{Ethical Concerns}
All of the data used in this paper is either our own or from publicly released and licensed sources. Our data is mainly aimed towards researchers and developers who wish to assess the qualities of word embedding models and gain some intuition for embedding model selection for downstream tasks. In particular, our conclusions would be suited for researchers working among the 29 functioning languages given in the MUSE library, which are heavily skewed towards Indo-European languages. Though we do not directly introduce novel NLP applications, we provide resources that may be useful in selecting technologies to deploy and informing the development of word embedding systems. 

Categorical modularity is intended to be an informational tool that sheds light on semantic representation of natural language information in computational word embeddings, and there are many aspects of its capabilities that can be improved upon, extended, or further explored. We would also like to emphasize that we have only tested our metric on three specific downstream tasks with basic downstream models, and these may not be representative of all NLP tasks in general. Categorical modularity also has not yet been shown to reveal information on representational harms inherent in word embedding spaces, so evidence of good downstream task performance should not be misconstrued as indicative of strong and beneficial performance across all NLP domains.

\subsection*{Environmental Impact}
We acknowledge the pressing threat of climate change and therefore record some statistics on the computational costs of our experiments. All of our experiments are run with a 13-inch 2019 MacBook Pro with a $1.7$ GHz Quad-Core Intel Core i7 processor running Python $3.8.3$ in Terminal Version $2.11$ on MacOS Big Sur Version $11.1$. For the English language, generating FastText embeddings for our 500 core words took $20.31$ seconds, generating the $500 \times 500$ $k$-NNM took  1 hour and 25.72 seconds, generating MUSE embeddings for the 500 words took $23.13$ seconds, and generating the $500 \times 500$ $k$-NNM took 7 minutes and $39.32$ seconds. For the downstream task of movie review sentiment analysis, it took $42.03$ seconds to generate FastText sentence embeddings for $10{,}000$ English reviews and 6 minutes and $38.32$ seconds to generate these embeddings with MUSE. It took $0.35$ seconds per review to translate from English to Spanish using the Google Translate API, and it took $2.6$ seconds to run 30 trials of the sentiment analysis task for English FastText using scikit-learn's LinearSVC. For the task of word similarity calculation, English FastText embeddings and 3-dimensional input data took $21.25$ seconds to generate for 500 word pairs, English MUSE-based embedding data took $33.47$ seconds to generate, and the word similarity task using scikit-learn's LinearRegression took $0.09$ seconds on the generated English FastText-based inputs. For bilingual lexicon induction, FastText English-Spanish embedding data took $55.87$ seconds to generate, MUSE English-Spanish embedding data took 27 minutes and $29.56$ seconds to generate, and the BLI task took a combined $4.35$ seconds for both directions of English-Spanish using FastText. All other tasks took less than one second per language/model pair.

\appendix
\section{Appendix: Full Results for Correlations of Categorical Modularities with Downstream Tasks}\label{sec:appa}
This section contains full tables of correlations of general categorical modularities and unsupervised cluster modularities with downstream tasks. As above, $\rho$ is overall correlation, $\rho_{ft}$ is the correlation within FastText, $\rho_m$ is the correlation within MUSE, and $\rho_s$ is the correlation within subs2vec. On notation: in the ``Model'' columns, $a, b$ represents the hyperparameters of Level $a$ Binder categories with $k = b$ neighbors, while ``C, $a$'' represents ``control'' unsupervised clusters with the hyperparameter of $k = a$ neighbors.

\subsection{Sentiment Analysis}

\begin{table}[H]
\centering
\begin{tabular}{crcrc} 
\hline
\textbf{Model} & $\mathbf{\rho}\;\;\;$ & $\mathbf{\rho_{ft}}$ & $\mathbf{\rho_m}\;\;$ & $\mathbf{\rho_s}$ \\[0.05cm]
\hline
1, 2 & 0.49 & 0.25 & 0.45 & 0.44 \\
1, 3 & 0.49 & 0.26 & 0.47 & 0.54 \\
1, 4 & 0.45 & 0.21 & 0.33 & 0.53 \\
2, 2 & 0.52 & 0.32 & 0.53 & 0.45 \\
2, 3 & 0.51 & 0.32 & 0.58 & 0.59 \\
2, 4 & 0.47 & 0.36 & 0.52 & 0.54 \\
3, 2 & 0.54 & 0.44 & 0.68 & 0.46 \\
3, 3 & 0.51 & 0.34 & 0.72 & 0.58 \\
3, 4 & 0.44 & 0.33 & 0.70 & 0.50 \\
C, 2 & 0.09 & 0.10 & 0.40 & 0.35 \\
C, 3 & $-0.21$ & 0.13 & $-0.04$ & 0.15 \\
C, 4 & $-0.25$ & 0.05  & $-0.24$ & 0.15 \\
\hline
\end{tabular}
\caption{Spearman correlations of categorical modularity with accuracy of IMDB sentiment analysis task.}
\label{apptable:1}
\end{table}


\begin{table}[H]
\centering
\begin{tabular}{crcrc} 
\hline
\textbf{Model} & $\mathbf{\rho}\;\;\;$ & $\mathbf{\rho_{ft}}$ & $\mathbf{\rho_m}\;\;$ & $\mathbf{\rho_s}$ \\[0.05cm]
\hline
1, 2 & 0.45 & 0.27 & 0.44 & 0.27 \\
1, 3 & 0.45 & 0.29 & 0.45 & 0.36 \\
1, 4 & 0.42 & 0.25 & 0.31 & 0.35 \\
2, 2 & 0.47 & 0.32 & 0.46 & 0.29 \\
2, 3 & 0.48 & 0.32 & 0.53 & 0.44 \\
2, 4 & 0.45 & 0.38 & 0.49 & 0.38 \\
3, 2 & 0.49 & 0.42 & 0.60 & 0.29 \\
3, 3 & 0.46 & 0.32 & 0.66 & 0.42 \\
3, 4 & 0.41 & 0.34 & 0.64 & 0.35 \\
C, 2 & 0.06 & 0.04 & 0.37 & 0.18 \\
C, 3 & $-0.18$ & 0.14 & $-0.08$ & 0.02 \\
C, 4 & $-0.24$ & 0.09  & $-0.31$ & 0.03 \\
\hline
\end{tabular}
\caption{Spearman correlations of categorical modularity with precision of IMDB sentiment analysis task.}
\label{apptable:2}
\end{table}

\subsection{Word Similarity}

\begin{table}[H]
\centering
\setstretch{0.97}
{\normalsize
\begin{tabular}{crcrc} 
\hline
\textbf{Model} & $\mathbf{\rho}\;\;\;$ & $\mathbf{\rho_{ft}}$ & $\mathbf{\rho_m}\;\;$ & $\mathbf{\rho_s}$ \\[0.05cm]
\hline
1, 2 & $-0.66$ & $-0.47$ & $-0.36$ & $-0.74$ \\
1, 3 & $-0.60$ & $-0.46$ & $-0.24$ & $-0.71$ \\
1, 4 & $-0.57$ & $-0.46$ & $-0.15$ & $-0.71$ \\
2, 2 & $-0.71$ & $-0.59$ & $-0.34$ & $-0.80$ \\
2, 3 & $-0.65$ & $-0.62$ & $-0.29$ & $-0.80$ \\
2, 4 & $-0.61$ & $-0.60$ & $-0.23$ & $-0.78$ \\
3, 2 & $-0.69$ & $-0.65$ & $-0.44$ & $-0.79$ \\
3, 3 & $-0.62$ & $-0.60$ & $-0.42$ & $-0.81$ \\
3, 4 & $-0.57$ & $-0.61$ & $-0.32$ & $-0.79$ \\
C, 2 & $-0.27$ & $-0.36$ & $-0.30$ & $-0.42$ \\
C, 3 & 0.00 & $-0.29$ & 0.20 & $-0.45$ \\
C, 4 & 0.04 & $-0.36$ & 0.32 & $-0.36$ \\
\hline
\end{tabular}
}
\caption{Correlations with word similarity mean MSE.}
\label{apptable:3}
\end{table}

\subsection{Bilingual Lexicon Induction}

\begin{table}[H]
\centering
\setstretch{0.97}
{\normalsize
\begin{tabular}{ccrrc} 
\hline
\textbf{Model} & $\rho$ & $\mathbf{\rho_{ft}}\;\;$ & $\mathbf{\rho_m}\;\;$ & $\mathbf{\rho_s}$ \\[0.05cm]
\hline
1, 2 & 0.41 & 0.15 & 0.33 & 0.68 \\
1, 3 & 0.33 & 0.07 & 0.24 & 0.64 \\
1, 4 & 0.29 & 0.05 & 0.16 & 0.64 \\
2, 2 & 0.50 & 0.34 & 0.42 & 0.77 \\
2, 3 & 0.41 & 0.24 & 0.37 & 0.74 \\
2, 4 & 0.38 & 0.22 & 0.30 & 0.73 \\
3, 2 & 0.55 & 0.39 & 0.53 & 0.76 \\
3, 3 & 0.49 & 0.35 & 0.47 & 0.75 \\
3, 4 & 0.45 & 0.31 & 0.37 & 0.74 \\
C, 2 & 0.35 & 0.19 & 0.44 & 0.55 \\
C, 3 & 0.09 & $-0.09$ & $-0.17$ & 0.46 \\
C, 4 & 0.07 & $-0.06$ & $-0.09$ & 0.37 \\
\hline
\end{tabular}
}
\caption{Correlations with mean cosine similarity on BLI from English.}
\label{apptable:4}
\end{table}

\begin{table}[H]
\centering
\setstretch{0.97}
{\normalsize
\begin{tabular}{ccrcc} 
\hline
\textbf{Model} & $\rho$ & $\mathbf{\rho_{ft}}\;\;$ & $\mathbf{\rho_m}$ & $\mathbf{\rho_s}$ \\[0.05cm]
\hline
1, 2 & 0.32 & 0.05 & 0.35 & 0.74 \\
1, 3 & 0.28 & $-0.05$ & 0.33 & 0.71 \\
1, 4 & 0.26 & $-0.09$ & 0.32 & 0.70 \\
2, 2 & 0.42 & 0.22 & 0.45 & 0.81 \\
2, 3 & 0.37 & 0.10 & 0.46 & 0.79 \\
2, 4 & 0.38 & 0.07 & 0.47 & 0.78 \\
3, 2 & 0.50 & 0.29 & 0.56 & 0.81 \\
3, 3 & 0.47 & 0.20 & 0.52 & 0.80 \\
3, 4 & 0.46 & 0.16 & 0.51 & 0.78 \\
C, 2 & 0.35 & 0.04 & 0.27 & 0.65 \\
C, 3 & 0.29 & $-0.21$ & 0.15 & 0.57 \\
C, 4 & 0.25 & $-0.13$ & 0.19 & 0.48 \\
\hline
\end{tabular}
}
\caption{Correlations with mean cosine similarity on BLI to English.}
\label{apptable:5}
\end{table}

\newpage

\section{Sizes of Embedding Models}\label{app:b}
For contextual reference, we summarize the sizes of each of the embedding models used in this paper.

\begin{table}[H]
\centering
\begin{tabular}{crrr} 
\hline
\textbf{Language} & \textbf{FastText}\, & \textbf{MUSE}\,\, & \textbf{subs2vec} \\[0.05cm]
\hline
Arabic	&	610{,}976	&	132{,}480	&	898{,}080	\\
Bulgarian	&	334{,}077	&	200{,}000	&	753{,}982	\\
Catalan	&	490{,}564	&	200{,}000	&	27{,}220	\\
Croatian	&	451{,}636	&	200{,}000	&	1{,}000{,}000	\\
Czech	&	627{,}840	&	200{,}000	&	1{,}000{,}000	\\
Danish	&	312{,}955	&	200{,}000	&	262{,}951	\\
Dutch	&	871{,}021	&	200{,}000	&	495{,}055	\\
English	&	1{,}000{,}000	&	200{,}000	&	1{,}000{,}000	\\
Estonian	&	329{,}986	&	200{,}000	&	357{,}632	\\
Finnish	&	730{,}482	&	200{,}000	&	842{,}787	\\
French	&	1{,}000{,}000	&	200{,}000	&	514{,}066	\\
Greek	&	306{,}448	&	200{,}000	&	859{,}548	\\
Hebrew	&	488{,}935	&	200{,}000	&	679{,}649	\\
Hungarian	&	793{,}865	&	200{,}000	&	1{,}000{,}000	\\
Indonesian	&	300{,}685	&	200{,}000	&	221{,}876	\\
Italian	&	871{,}052	&	200{,}000	&	597{,}058	\\
Macedonian	&	176{,}946	&	176{,}947	&	132{,}238	\\
Norwegian	&	515{,}787	&	200{,}000	&	179{,}069	\\
Polish	&	1{,}000{,}000	&	200{,}000	&	1{,}000{,}000	\\
Portuguese	&	592{,}107	&	200{,}000	&	505{,}535	\\
Romanian	&	354{,}323	&	200{,}000	&	964{,}079	\\
Russian	&	1{,}000{,}000	&	200{,}000	&	802{,}112	\\
Slovak	&	316{,}097	&	200{,}000	&	330{,}354	\\
Slovenian	&	281{,}822	&	200{,}000	&	517{,}625	\\
Spanish	&	985{,}666	&	200{,}000	&	883{,}541	\\
Swedish	&	1{,}000{,}000	&	200{,}000	&	325{,}033	\\
Turkish	&	416{,}050	&	200{,}000	&	1{,}000{,}000	\\
Ukrainian	&	912{,}457	&	200{,}000	&	80{,}123	\\
Vietnamese	&	292{,}167	&	200{,}000	&	80{,}216	\\
\hline
\end{tabular}
\caption{Number of vectors in each (language, model) pair.}
\label{apptable:6}
\end{table}

\end{document}